# An Add-On for Empowering Google Forms to be an Automatic Question Generator in Online Assessments

**Pornpat Sirithumgul,[a] Pimpaka Prasertsilp,[b] Lorne Olfman[c]**

[a]Department of Computer Engineering, Rajamangala University of Technology Phra Nakhon, Thailand.

[b]School of Science and Technology, Sukhothai Thammathirat Open University, Thailand.

[c]Center for Information Systems and Technology, Claremont Graduate University, USA.

Authors Emails: pornpat.s@rmutp.ac.th. pimpaka.pra@stou.ac.th. lorne.olfman@cgu.edu.

ABSTRACT: This research suggests an add-on to empower Google Forms to be an automatic machine for generating multiple-choice questions (MCQs) used in online assessments. In this paper, we elaborate an add-on design mainly comprising question-formulating software and data storage. The algorithm as an intellectual mechanism of this software can produce MCQs at an analytical level. In an experiment, we found the MCQs could assess levels of students' knowledge comparably with those generated by human experts. This add-on can be applied generally to formulate MCQs for any rational concepts. With no effort from an instructor at runtime, the add-on can transform a few data instances describing rational concepts to be variety sets of MCQs.

KEYWORDS: Third-party software, ontology-based algorithm, automatic question generation, online knowledge assessment, online learning technology.

## 1. INTRODUCTION

Google Forms[1] is popularly used to create surveys and online exams because of its feature for arranging questions presentable for survey respondents and online test-takers [1]. That users can simply generate reports in multiple views from questionnaire responses is another important reason why Google Forms is more popular than other online questionnaire tools [2].

Despite several benefits of Google Forms, we argue this tool is not very practical for instructors to make online exams. In online assessments, instructors have serious concerns about fraudulent activities [3 - 6], and try to avoid such activities by producing multiple sets of questions before dispersing them to online students. The time an instructor spends on preparing an online exam is therefore much greater than that spent on a face-to-face exam. Also, multiple question sets cooperatively used in one exam may not be consistent, and possibly reduce the overall reliability of the exam.

An automatic question generator could be an alternative for reducing instructors' workloads, and preserving consistency of exam questions. Several prior research studies [7 – 17] suggested methods and software algorithms for constructing automatic question generators that truly succeeded in formulating questions that are readable and understandable by humans. A limitation, however, is the fact that these automatic question generators [7 – 17] were not evaluated in an actual academic setting, and their level of success as expert systems for assessing students' knowledge was not reported.

Many research studies [18 – 39] are very similar to this current research as they proposed automatic tools for generating multiple-choice questions (MCQs) in academia. The tools proposed in [18 – 22] were applied in linguistics, and those proposed in [23 – 39] were applied in scientific subjects. However, these studies have not shown whether topics/concepts and cognitive levels (e.g., remembering, understanding, analyzing) [40] presented in the same set questions are consistent.

This research is aimed to reduce the limitations of the above-noted studies by proposing an add-on with the following features:
1) **Compatibility with Google Forms.** This add-on compels Google Forms to be a MCQs generator, and yet it preserves the original modules of Google Forms making MCQs presentable to students, and making students' responses visualizable to instructors.
2) **Consistency of automated MCQs.** This add-on guarantees a consistency of the generated MCQs in terms of a topic/concept asked in the same set of questions as well as a cognitive level that the MCQs would be able to measure.
3) **Minimum upfront effort required.** The add-on requires less effort for feeding data instances to the system. No additional upfront effort or technical knowledge is required to deal with a system interface or complicated data used to formulate MCQs.

## 2. ALGORITHM AND ARCHITECTURAL DESIGN

The add-on is a system comprising four main constructs – data storage, executable software, spreadsheets, and Google Forms. *Data storage* is used to collect data instances for constructing MCQs. *Executable software* is used to formulate MCQs from data instances retrieved from the data storage. An output from the software is a collection of MCQs collected in a spreadsheet ready to be uploaded to Google Forms. Google Forms in this system is used as a user interface for presenting MCQs, and receiving users' responses.

*The key idea* of formulating MCQs by the proposed add-on is to combine relational data instances describing relative concepts in the same domain. Particularly in this research, we focus on 5 concepts in the software testing domain. These concepts include Python code, flow graphs, and three methods for testing software – *'path coverage'*, *'branch coverage'*, and *'statement coverage'*. Figures 1 - 2 show sample data instances of the five relative concepts in the data storage. The flow graph in Figure 1 (b) represents Python code in Figure 1 (a). This flow graph is used as an input for calculating the number of test cases by using *'path coverage'*, *'branch coverage'*, and *'statement coverage'* testing methods. As presented in Figure 2, the three numbers of test cases – 3, 2, and 1 are collected in the data storage.

---

[1] Google Forms (in Google Workspace) is a free online application among others, e.g., Google Docs, Google Sheets, Google Slides, and Google Drive (https://workspace.google.com, accessed on March 13, 2021).

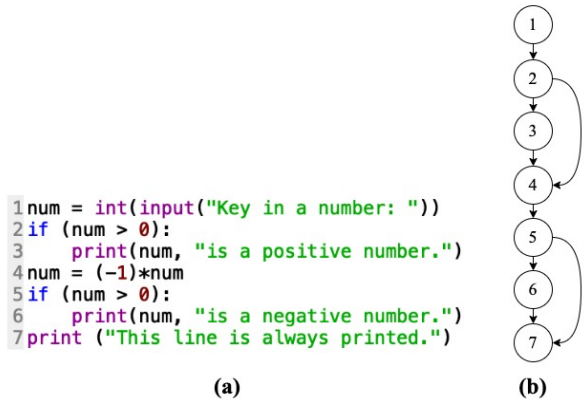

Figure 1. Python Code (a) and Flow Graph (b)

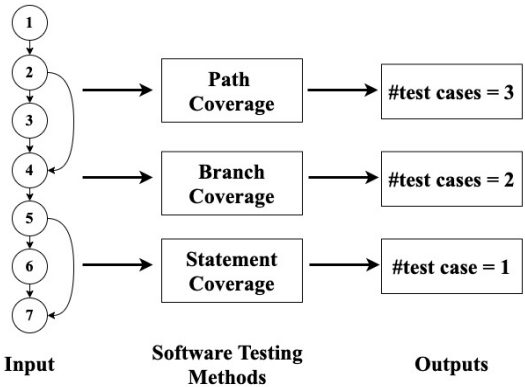

Figure 2. The Number of Test Cases Calculated by Three Software Testing Methods

*Executable Software* has an important role in formulating MCQs from data instances in the data storage. A MCQ comprises one question phrase, and four choices – one is a correct answer, and the others are distractors. Pseudocode in Figures 3 - 4 presents how this software formulates question phrases, and multiple choices.

*A question phrase* (see Figure 3) is composed from Python code or a flow graph combined with a statement that encourages students to use a software testing method to find Python code/a flow graph in an answer choice having the number of test cases equal to the Python code/flow graph in the question phrase.

The variable *'question_type'* in the pseudocode for constructing question phrases and multiple choices (see Figures 3 - 4) is used to decide how to combine Python code and flow graphs in a question phrase and multiple choices. If the variable is set to 1, flow graphs would be used both in the question phrase and multiple choices. If the variable is set to 3, Python code would be used both in the question phrase and multiple choices. However, if the variable is set to 2 or 4, a question phrase and multiple choices would be different – the former would use a flow graph in the question phrase, and Python code in four choices, and the latter would be in the other way around. The variable *'method'* (see Figures 3 – 4) is used to specify a software testing method – *'path coverage'*, *'branch coverage'*, or *'statement coverage'* being asked in the question, and the variable *'#test_cases'* (see Figure 4) is used to identify a number of test cases of Python code or a flow graph in a correct answer choice; the other three distractors must contain Python code or flow graphs having a different number of test cases.

---

**Pseudocode 1:** Question Phrase Formulation

1   **Function** Question_Phrase (**String** *method*, **Integer** *question_type*):
    /** **Inputs:**
    1) The variable *method* identifies one of the software testing methods in the set {*'path coverage'*, *'branch coverage'*, *'statement coverage'*}
    2) If the variable *question_type* is 1 or 2, a *flow graph* is used in the question phrase; otherwise, *Python code* is instead used.

    **Output:** A question phrase of a multiple-choice question **/
    /** **Retrieving data from the database** **/
2       **Select** Object_list
3       **From** Data_collection
4       **Where** Calculation_method = *method*

5       *Data_object* ← Randomly select from *Object_list*
6       **If** (question_type = 1 **OR** question_type = 2) **Then**
7           *Flow_graph* = *Data_object*.flow_graph
8           #*test_cases* = Data_object.test_cases
9           **Construct** → Question Phrase *(Flow_graph)*
10          **Call** Function Multiple_Choices (*method*, *question_type*, *#test_cases*)
11      **Else if** (question_type = 3 **OR** question_type = 4) **Then**
12          *Code* = Data_object.code
13          #*test_cases* = Data_object.test_cases
14          **Construct** → Question Phrase *(Code)*
15          **Call** Function Multiple_Choices (*method*, *question_type*, *#test_cases*)

Figure 3. Pseudocode for Formulating Question Phrases

---

**Pseudocode 2:** Multiple-choice Formulation

1   **Function** Multiple_Choices (**String** *method*, **Integer** *question_type*, **Integer** *#test_cases*):
    /** **Inputs:**
    1) The variable *method* identifies one of the software testing methods in the set {*'path coverage'*, *'branch coverage'*, *'statement coverage'*}
    2) If the variable *question_type* is 1 or 4, *flow graphs* are used in multiple choices; otherwise, four pieces of *Python code* are instead used.
    3) The variable *#test_cases* contains the number of test cases as calculated by the software testing method.

    **Output:** A set of four answer choices**/
    /** **Retrieving data from the database** **/
2       **Select** Object_list
3       **From** Data_collection
4       **Where** Calculation_method = *method*

5       **While** (Four choices have *not* yet been set) {
6           *Data_object* ← Randomly select from *Object_list*
7           **If** (question_type = 1 **OR** question_type = 4) **Then**
8               *Flow_graph* = *Data_object*.flow_graph
9               **If** (*#test_cases* is *equal* to Data_object.test_cases) **Then**
10                  **Construct** → Correct Answer Choice *(Flow_graph)*
11              **Else if** (*#test_cases* is *not* equal to Data_object.test_cases) **Then**
12                  **Construct** → Distractor *(Flow_graph)*
13          **Else if** (question_type = 2 **OR** question_type = 3) **Then**
14              *Code* = Data_object.code
15              **If** (*#test_cases* is *equal* to Data_object.test_cases) **Then**
16                  **Construct** → Correct Answer Choice *(Code)*
17              **Else if** (*#test_cases* is *not* equal to Data_object.test_cases) **Then**
18                  **Construct** → Distractor *(Code)*

Figure 4. Pseudocode for Formulating Answer Choices

Outputs from the software are MCQs in four types. Figures 5 - 6 are examples of the MCQs in four types presented in Google Forms.

1) **Type 1: Flow graphs in the question phrase and four answer choices.** Given a software testing method, students would ask to map two flow graphs sharing the same number of test cases.
2) **Type 2: Flow graph in the question phase and Python code in four answer choices.** Given a software testing method,

students would ask to map the flow graph with Python code in an answer choice having the same number of test cases.

3) **Type 3: Python code in the question phrase and four answer choices.** Given a software testing method, students would ask to map two pieces of Python code having the same number of test cases.

4) **Type 4: Python code in the question phrase and flow graphs in four answer choices.** Given a software testing method, students would ask to map the Python code with a flow graph in an answer choice having the same number of test cases.

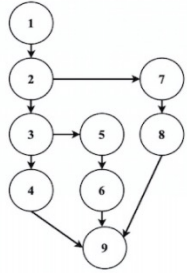
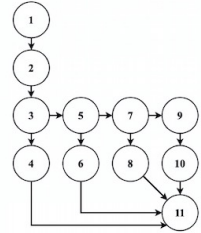

(a)													(b)

Figure 5. MCQs Type 1 (a) and Type 2 (b)

(a)													(b)

Figure 6. MCQs Type 3 (a) and Type 4 (b)

## 3. EXPERIMENTAL DESIGN

An experiment was conducted to compare quality of MCQs generated by the add-on with those composed by human experts. As presented in the experimental procedure (see Figure 7), 60 MCQs were generated by the add-on, and divided into 5 sets, each of which had 12 MCQs. These 12 MCQs were in 3 subgroups separately asking about path coverage, branch coverage, and statement coverage. Each subgroup contained 4 MCQs in Types 1 – 4.

A participant in this experiment was requested to answer one set of 12 MCQs and one set of 12 baseline questions. These baseline questions were retrieved from two tutorial webpages[2,3] posting questions about software testing including the three concepts – path coverage, branch coverage, and statement coverage employed in this baseline set.

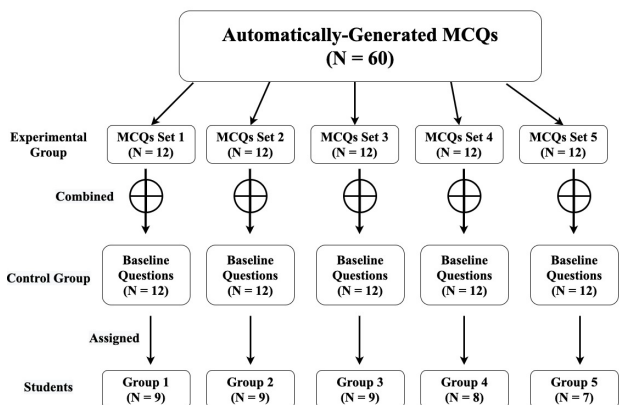

Figure 7. Experimental Procedure

Figure 8 is an example of the baseline questions asking students to use a branch coverage testing method to calculate the number of test cases from the given pseudocode.

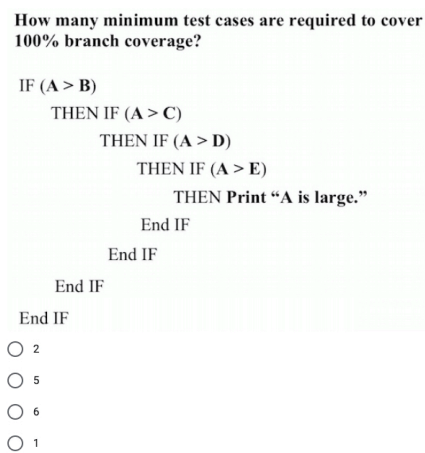

Figure 8. An Example of Baseline Questions

Forty-two participants were randomly assigned to 5 groups. Groups 1 – 3 had 9 participants, Group 4 had 8, and Group 5 had 7. Participants in Groups 1 – 5 were assigned to answer the same set of baseline questions, and different sets of automated MCQs.

---

[2] *How to calculate Statement, Branch/Decision and Path Coverage for ISTQB Exam purpose.* Retrieved from https://www.istqb.guru/how-to-calculate-statement-branchdecision-and-path-coverage-for-istqb-exam-purpose/ on February 3, 2021.
[3] *Testing Manual Automation.* Retrieved from

The participants in this experiment were 42 junior students – 30 were male and 12 were female – in the department of Computer Engineering at a technology university in Thailand. Before this experiment, the students had passed two computer programming courses – one was basic programming using C, and the other was object-oriented programming using Python. Two weeks before participating in this experiment, the students learned software testing in a Software Engineering course. Literally, the students spent 3 hours learning the black-box testing approach, and another 3 hours learning the white-box testing approach including the three concepts – path coverage, branch coverage, and statement coverage used in this experiment. The students spent around 60 to 90 minutes answering the MCQs and baseline questions.

The outputs from this experiment are students' answers to the MCQs and to the baseline questions. The answers from the two groups of questions are compared and analyzed in the next section.

## 4. EXPERIMENTAL RESULTS

The participants' answers to the MCQs and baseline questions are compared and analyzed in this section. We found that on average students could correctly answer 7.40 MCQs ($\mu$ = 7.40, $\sigma$ = 1.88), and 7.81 baseline questions ($\mu$ = 7.81, $\sigma$ = 2.03). Correct answers to MCQs were in the range of 1 to 12, and correct answers to the baseline questions were between 3 and 12. As presented in the frequency distribution table (Table 1), the mode for both sets of questions was 8 out of 12 correct answers. By considering individual students' scores, we found a high correlation between the number of correct answers to MCQs and to baseline questions. The statistical data $r(40) = 0.86$, p-value < .00001 (see Figure 9) clearly show an agreement of student performance measured by MCQs and baseline questions.

Table 1. Frequency Distribution Table

| #Correct Answers | #Participants Answering MCQs | #Participants Answering Baseline Questions |
|---|---|---|
| 1 | 1 (2.38%) | 0 (0.00%) |
| 2 | 0 (0.00%) | 0 (0.00%) |
| 3 | 1 (2.38%) | 1 (2.38%) |
| 4 | 1 (2.38%) | 2 (4.76%) |
| 5 | 2 (4.76%) | 2 (4.76%) |
| 6 | 5 (11.90%) | 5 (11.90%) |
| 7 | 7 (16.67%) | 8 (19.05%) |
| 8* | **15 (35.71%)** | **10 (23.81%)** |
| 9 | 8 (19.05%) | 4 (9.52%) |
| 10 | 1 (2.38%) | 6 (14.29%) |
| 11 | 0 (0.00%) | 3 (7.14%) |
| 12 | 1 (2.38%) | 1 (2.38%) |
|  | N = 42 (100%) | N = 42 (100%) |

https://testsoftwarefaq.blogspot.com/2013/10/examples-of-statement-branchdecision.html on February 3, 2021.

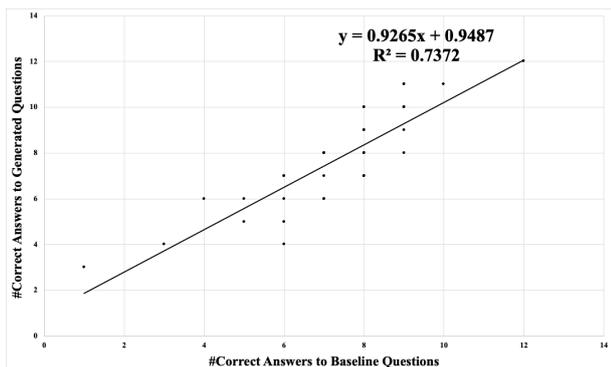
Figure 9. Correlation between Correct Answers to MCQs and Baseline Questions

We can draw a conclusion based on this experimental result that the add-on proposed in this research can produce MCQs for which quality is as high as those produced by software-testing domain experts. This add-on is therefore worth using to substitute humans' efforts in producing questions for assessing knowledge in this domain.

Tables 2 - 4 elaborate statistical data for the five sets of questions that were combined between MCQs and baseline questions assigned to five groups of participants. We found R scores showing correlations between correct answers to MCQs and baseline questions of all sets were about equal, i.e., 0.87, 0.88, 0.85, 0.80, and 0.87. These correlations reveal the quality of the add-on in producing consistent MCQs.

Table 2. Statistical Data of Correct Answers to Questions Assigned to Students Groups 1 – 2

| Student Group 1 (N=9) | | Student Group 2 (N=9) | |
|---|---|---|---|
| MCQs (N = 12) | Baseline Questions (N = 12) | MCQs (N = 12) | Baseline Questions (N = 12) |
| μ = 5.11 σ = 2.08 | μ = 5.78 σ = 1.62 | μ = 8.22 σ = 1.69 | μ = 8.00 σ = 2.11 |
| R scores = 0.87, P-values < .00001 | | R scores = 0.88, P-values < .00001 | |

Table 1. Statistical Data of Correct Answers to Questions Assigned to Students Groups 3 – 4

| Student Group 3 (N=9) | | Student Group 4 (N=8) | |
|---|---|---|---|
| MCQs (N = 12) | Baseline Questions (N = 12) | MCQs (N = 12) | Baseline Questions (N = 12) |
| μ = 8.44 σ = 0.83 | μ = 9.33 σ = 1.06 | μ = 7.63 σ = 1.11 | μ = 8.13 σ = 1.45 |
| R scores = 0.85, P-values < .00001 | | R scores = 0.80, P-values < .00001 | |

Table 4. Statistical Data of Correct Answers to Questions Assigned to Group 5

| Group 5 (N=7) | |
|---|---|
| MCQs (N = 12) | Baseline Questions (N = 12) |
| μ = 7.71 σ = 0.70 | μ = 7.86 σ = 1.81 |
| R scores = 0.87, P-values < .00001 | |

We found the distractors of the MCQs were effective. Referring to the previous research [41 – 48], an effective distractor means one chosen by greater than 5 percent of the number of students answering the question containing the distractor.

MCQs in sets 1 – 5 totally contain 106 effective distractors. Table 5 presents the number of students choosing distractors in the five MCQ sets. The distractors in MCQ sets 1 – 3 were chosen by 1 - 6 students (10.0 – 60.0 %). The distractors in MCQ set 4 were chosen by 1 – 4 students (12.5 – 50.0 %), and the distractors in MCQ set 5 were chosen by 1 – 4 students (14.3 – 57.1 %).

Table 5. Effective Distractors in MCQ Sets 1 - 5

| MCQ Set | #Students (%) | #Distractors |
|---|---|---|
| 1 | 10.0 – 60.0% | 27 |
| 2 | 10.0 – 50.0% | 17 |
| 3 | 10.0 – 50.0% | 20 |
| 4 | 12.5 – 50.0% | 20 |
| 5 | 14.3 – 57.1% | 22 |
| | | Total = 106 |

Quality of distractors reflects the overall quality of MCQs. The previous research [41 – 48] suggests that an MCQ is effective if there is at least one effective distractor in the multiple-choice set. Table 6 presents the number of effective distractors occurring in the MCQ sets. All 12 MCQs in set 1 are effective – there are 2 MCQs containing one effective distractor, 5 MCQs containing two effective distractors, and 5 MCQs containing three effective distractors. There are 4 MCQs in sets 2 – 4 not containing effective distractors. All students assigned these 4 MCQs made the correct answer choices, and ignored all distractors. Based on a consideration of effective distractors, we conclude that 56 out of 60 MCQs (about 93 %) are effective, and can be suitably used as a knowledge measure.

Table 6. The Number of Effective Distractors in MCQ Sets 1 – 5

| | #Effective Distractors | | | | Total MCQs |
|---|---|---|---|---|---|
| | Three | Two | One | Zero | |
| Set 1 | 5 | 5 | 2 | 0 | 12 |
| Set 2 | 1 | 4 | 6 | 1 | 12 |
| Set 3 | 3 | 3 | 5 | 1 | 12 |
| Set 4 | 2 | 5 | 4 | 1 | 12 |
| Set 5 | 3 | 5 | 3 | 1 | 12 |
| Total | 14 | 22 | 20 | 4 | 60 |

## 5. DISCUSSIONS

In this section, we discuss compatibility between the proposed add-on and Google Forms in the Google Workspace, consistency of a concept and a cognitive level presented in the automatically generated MCQs, generalizability of the add-on if applied to generate MCQs in other knowledge domains, and effort reduction of instructors in generating MCQs and preparing upfront data.

*5.1 Compatibility between the add-on and Google Workspace.*

Google Forms can compatibly work with other applications on Google Workspace, i.e., Google Docs, Google Sheets and Google Slides. Data from Google Forms' questionnaires can be exported to Google Sheets, and vice versa, data from Google Docs, Google Sheets and Google Slides can be exported to Google Forms. However, Google Forms does not have an embedded function for importing data from such companion applications. Users need to download an add-on called *Form*

Builder[4], and use the add-on to transfer data to Google Forms. Figure 10 presents how the add-on presented in this research works with the Google Workspace.

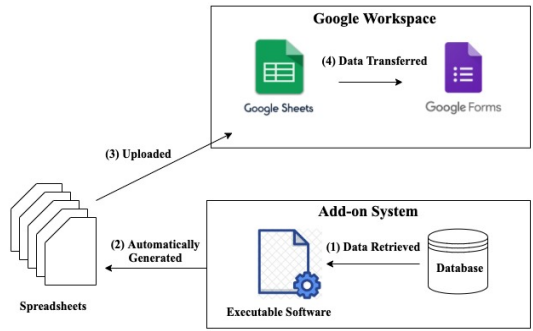

Figure 10. Compatibility between the Proposed Add-on and Google Workspace

The add-on (see Figure 10) comprises executable software and a database. At step (1) the executable software retrieves data from the database to generate MCQs collected in spreadsheets at step (2). These spreadsheets are in a form ready to upload to Google Workspace at step (3). We recommend instructors upload the spreadsheets to Google Sheets to keep the structure of data categorized in the 5 main columns of spreadsheets including one column of question phrases, one column of correct answers, and three columns of distractors. At step (4), the instructors use Form Builder to transfer data from Google Sheets to Google Forms.

*5.2 Consistency of MCQs generated by the add-on.*

The MCQs generated by the add-on are consistent in terms of concepts and a cognitive level. As controlled by data in the database, the MCQs cover three main concepts – path coverage, branch coverage, and statement coverage in the software testing domain.

Additionally, the MCQs in this research can be aligned with analytical questions, as shown in Ragonis [49] and Sirithumgul et al. [50], where students are encouraged to first analyze a problem and a solution in the question phrase, and then find an alternate solution from the given choices. Likewise, MCQs of this research encouraged students to first analyze software in the form of programming code and flow graphs, and then choose one from the four choices sharing the same characteristics with one in the question phrase.

*5.3 Generalizability of the add-on.*

An add-on would fit with one knowledge domain. However, the idea for constructing an add-on is generalizable, especially for a knowledge domain containing relational concepts. We can make an inductive conclusion based on the experiment reported here that in constructing an add-on, an ontology of the domain should be specified first to describe concepts and concepts' relations in the real world. These concepts and concepts' relations are later derived to be data instances in the database of the add-on.

Figure 11 presents an ontology for constructing the add-on of this research. The three concepts *'Python Code'*, *'Flow Graph'* and *'Software Testing Methods'* have mutual relations meaning that Python code is *'represented by'* flow graphs, and flow graphs would be *'used by'* software testing methods. As specified in the ontology, software testing methods include the three instance concepts – *'Path Coverage'*, 'Branch Coverage', and 'Statement Coverage'. Relations 'is-a' show 'Software Testing Methods' is a generic idea specified by the three instance concepts.

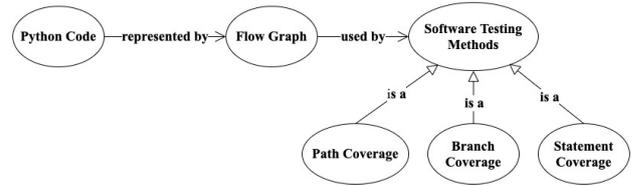

Figure 11. Concepts and Concepts' Relations

Figure 12 represents data instances in the database of the add-on. Data instances $c_1, c_2, c_3, \ldots c_n$ in entity *PYTHON CODE* represent Python code segments, data instances $g_1, g_2, g_3, \ldots g_n$ in entity *FLOW GRAPH* represent flow graphs as representatives of Python code segments, and data instances $m_1, m_2, m_3$ in entity *SOFTWARE TESTING METHODS* represent three specific testing methods – path coverage, branch coverage, and statement coverage. The relations between the entities *PYTHON CODE* and *FLOW GRAPH* are one-to-one, and the relations between the entities *FLOW GRAPH* and *SOFTWARE TESTING METHODS* are many-to-many. An associative entity between *FLOW GRAPH* and *SOFTWARE TESTING METHODS* is derived in this case to collect data instances about test cases as outputs of applying software testing methods on flow graphs.

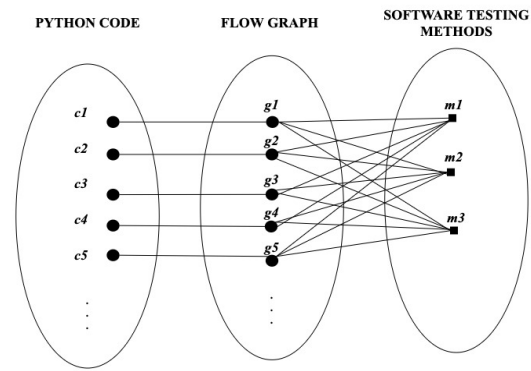

Figure 12. Data Instances in the Database

*5.4 Effort reduction of instructors in generating MCQs and preparing upfront data.*

The add-on proposed in this research requires minimum effort by an instructor to upload a few data instances to the database at design time to automatically generate MCQs at runtime. In the experiment reported here, we uploaded Python code, flow graphs, and numeric data identifying the number of test cases in the database. The algorithm of the executable software in this add-on uses a data combination technique in producing a tremendous number of MCQs from very few data instances. Suppose there are N data instances in the database and M out of N instances represent each other, in generating an MCQ, *two* out of M instances would be randomly selected – one would be used in a question phase and the other would be used in a correct answer choice; *three* out of N-M instances would be randomly selected and used as distractors. Totally, the number of different MCQs generated from this dataset would be the combination $\binom{M}{1}\binom{M-1}{1}\binom{N-M}{3}$.

In the experiment of this research, we uploaded 20 Python code segments, 20 flow graphs representing the Python code,

---

[4] Form Builder. https://formbuilder.jivrus.com/. Accessed on March 13, 2021.

and 60 numerous records representing the number of test cases from applying software testing methods on the flow graphs. In this dataset, 12 groups of Python code and 12 groups of flow graphs shared the same number of test cases. The same-group Python code and flow graphs were used to represent each other in the question phases and correct answers of MCQs. Totally, 277,984 different MCQs were generated from the small dataset in the experiment. We found 69,496 MCQs in each of Types 1 – 4 equally, and if considering concepts asked in the MCQs, we found 120,176 MCQs were about path coverage, 54,480 MCQs were about branch coverage, and 103,328 MCQs were about statement coverage.

6. CONCLUSION

This research proposes an add-on as a multiple-choice question generator working compatibly with Google Forms. This add-on has an important feature in consistently producing analytical questions for assessing knowledge in a given domain. Based on the experiment described in this paper, we found questions generated from the add-on were about as effective as those formulated by human experts. We also found this add-on could be generalized to formulate questions in any knowledge domain for which rational concepts were well defined. Most importantly, the data combination technique of this add-on can significantly reduce instructors' workloads. The instructors would receive a number of varied questions only by feeding a few data instances into the system.

7. ACKNOWLEDGEMENT

This research was jointly supported by National Science and Technology Development Agency (NSTDA) of Thailand [grant code FDA-CO-2562-8930-TH]; the Educational Exchange Program under the Collaboration between National Research Council of Thailand (NRCT) and Indian Council of Social Science Research (ICSSR); and Rajamangala University of Technology Phra Nakhon, Thailand.